\documentclass[letterpaper]{article} 
\usepackage{aaai25}  
\usepackage{times}  
\usepackage{helvet}  
\usepackage{courier}  
\usepackage[hyphens]{url}  
\usepackage{graphicx} 
\usepackage{natbib}  
\usepackage{caption} 
\usepackage{algorithm}
\usepackage{algorithmic}

\usepackage{newfloat}
\usepackage{listings}
\DeclareCaptionStyle{ruled}{labelfont=normalfont,labelsep=colon,strut=off} 
\floatstyle{ruled}
\newfloat{listing}{tb}{lst}{}
\floatname{listing}{Listing}
\usepackage{color,xcolor}
\usepackage{epsfig}
\usepackage{graphicx}

\usepackage{adjustbox}
\usepackage{array}
\usepackage{booktabs}
\usepackage{colortbl}
\usepackage{hhline}
\usepackage{multirow}
\usepackage{subcaption} 
\usepackage{floatflt}

\usepackage{amsmath,amsfonts,amssymb}
\usepackage{mathtools}  
\usepackage{bm}
\usepackage{nicefrac}
\usepackage{microtype}

\usepackage{xspace}
\usepackage[hidelinks]{hyperref}

\usepackage{amsmath,amsfonts,bm}

\def\eqref#1{equation~\ref{#1}}

\def\1{\bm{1}}

\DeclareMathAlphabet{\mathsfit}{\encodingdefault}{\sfdefault}{m}{sl}
\SetMathAlphabet{\mathsfit}{bold}{\encodingdefault}{\sfdefault}{bx}{n}

\def\gA{{\mathcal{A}}}

\def\gG{{\mathcal{G}}}

\def\gS{{\mathcal{S}}}
\def\gT{{\mathcal{T}}}
\def\gU{{\mathcal{U}}}

\newcommand{\R}{\mathbb{R}}

\newcolumntype{L}[1]{>{\raggedright\let\newline\\\arraybackslash\hspace{0pt}}m{#1}}
\newcolumntype{C}[1]{>{\centering\let\newline\\\arraybackslash\hspace{0pt}}m{#1}}
\newcolumntype{R}[1]{>{\raggedleft\let\newline\\\arraybackslash\hspace{0pt}}m{#1}}

\newcommand{\fig}[1]{Fig.~\ref{#1}}

\newcommand{\ignore}[1]{}

\makeatletter
\DeclareRobustCommand\onedot{\futurelet\@let@token\@onedot}
\def\@onedot{\ifx\@let@token.\else.\null\fi\xspace}

\def\eg{e.g\onedot} 
\def\ie{i.e\onedot}

\makeatother

\definecolor{MyDarkBlue}{rgb}{0,0.08,1}
\definecolor{MyDarkGreen}{rgb}{0.02,0.6,0.02}
\definecolor{MyDarkRed}{rgb}{0.8,0.02,0.02}
\definecolor{MyDarkOrange}{rgb}{0.40,0.2,0.02}
\definecolor{MyPurple}{RGB}{111,0,255}
\definecolor{MyRed}{rgb}{1.0,0.0,0.0}
\definecolor{MyGold}{rgb}{0.75,0.6,0.12}
\definecolor{MyDarkgray}{rgb}{0.66, 0.66, 0.66}

\usepackage{pifont} 

\newcommand{\benchmark}{HandMeThat\xspace}

\newcommand{\model}{FISER\xspace}

\def\bR{\mathbb{R}}

\newcommand{\xhdr}[1]{\noindent\textbf{#1}}

\newcommand{\revise}[1]{#1}

\nocopyright
\title{Infer Human's Intentions Before Following Natural Language Instructions}
\author{
    Yanming Wan\textsuperscript{\rm 1}, Yue Wu\textsuperscript{\rm 1}, Yiping Wang\textsuperscript{\rm 1}, Jiayuan Mao\textsuperscript{\rm 2}\equalcontrib, Natasha Jaques\textsuperscript{\rm 1}\equalcontrib
}
\affiliations {
    \textsuperscript{\rm 1}University of Washington, Seattle, WA 98195\\
    \textsuperscript{\rm 2}MIT CSAIL, Cambridge, MA 02139\\
    \{ymwan, nj\}@cs.washington.edu,
    jiayuanm@mit.edu
}

\usepackage{bibentry}

\begin{document}

\maketitle

\begin{abstract}
    For AI agents to be helpful to humans, they should be able to follow natural language instructions to complete everyday cooperative tasks in human environments. However, real human instructions inherently possess ambiguity, because the human speakers assume sufficient prior knowledge about their hidden goals and intentions. Standard language grounding and planning methods fail to address such ambiguities because they do not model human internal goals as additional partially observable factors in the environment. We propose a new framework, 
    \revise{Follow Instructions with Social and Embodied Reasoning  (\model)\footnote{Project website: \url{https://sites.google.com/view/fiser-hmt/}}},
    aiming for better natural language instruction following in collaborative embodied tasks. \revise{Our framework} makes explicit inferences about human goals and intentions as intermediate reasoning steps. 
    We implement a set of Transformer-based models and evaluate them over a challenging benchmark, HandMeThat. We empirically demonstrate that using social reasoning to explicitly infer human intentions before making action plans surpasses purely end-to-end approaches. We also compare our implementation with strong baselines, including Chain of Thought prompting on the largest available pre-trained language models, and find that \revise{FISER} provides better performance on the embodied social reasoning tasks under investigation, reaching the state-of-the-art on HandMeThat. 
\end{abstract}
\vspace{0.5em}
\section{Introduction}
\vspace{0.5em}
Building AI assistants that can interact with people in a shared environment and follow their instructions would unlock assistive robotics and free up domestic labor. Toward this broad goal, we need to address the problem of ``translating'' realistic natural language instructions into actions executable by robots.
The conventional way that people formulate this problem is grounded language learning, which aims at mapping abstract natural language phrases to concretely executable actions.  
However, these approaches miss an important component of many human-robot collaborative tasks, which is that the language humans tend to use in everyday scenarios is inherently ambiguous. Human speakers assume that listeners possess prior knowledge, leading them to omit certain information for efficiency~\cite{grice1975logic,sperber1986relevance,clark1996using,dennett1987intentional,gergely1995taking}.
Resolving this ambiguity depends on leveraging other sources of information (\eg, human internal goals and historical actions) that are partially observable to the robot.

\begin{figure*}[t]
    \centering
    \includegraphics[width=\textwidth]{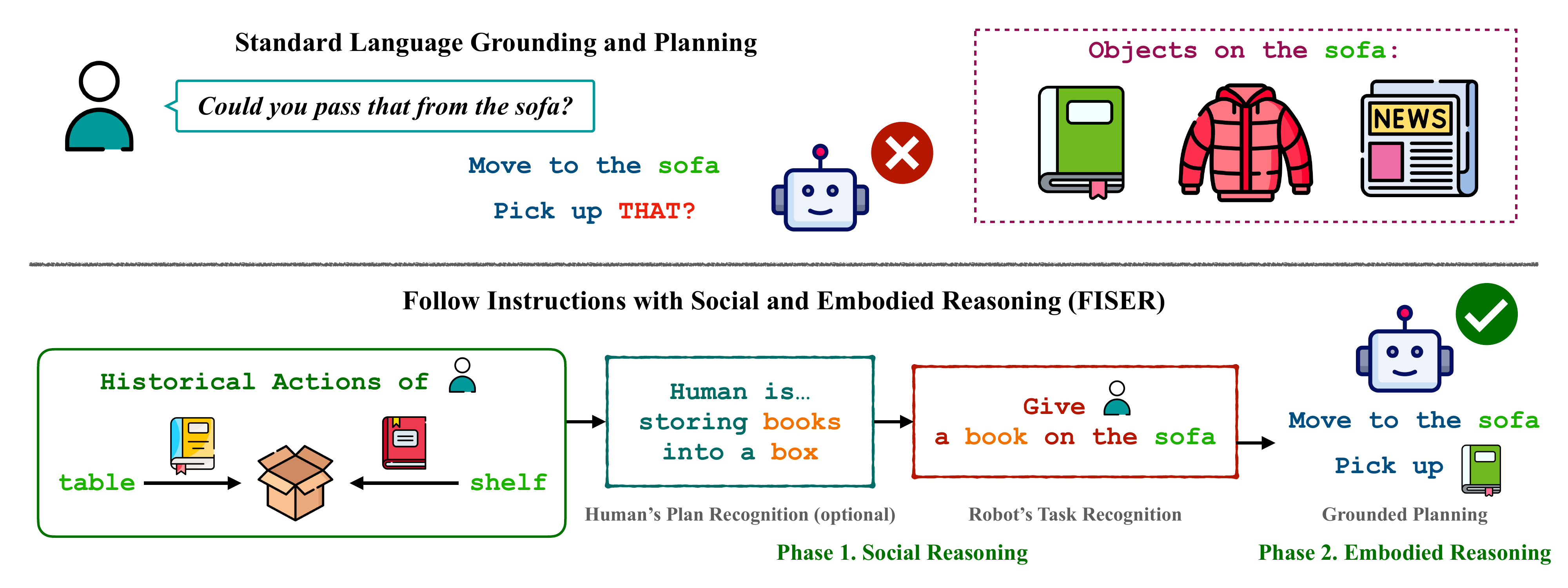}
    \caption{An example scenario where the human's natural language instruction (``\textit{Could you pass that from the sofa?}'') is inherently ambiguous. Standard language grounding and planning methods fail to resolve ambiguity. We propose \revise{\model}, which explicitly reasons about human's internal intentions as intermediate steps. \revise{The robot disamiguates the instruction into a concrete robot-understandable task in the social reasoning phase (Phase 1), and then accomplishes the grounded planning in the embodied reasoning phase (Phase 2). We further propose an optional enhancement to Phase 1 by explicitly recognize the human's overall plan first, and then infer what the human wants the robot to do.} 
}
    \label{fig:teaser}
\end{figure*}

Consider the example shown in \fig{fig:teaser}, where a human is tidying up a room. In the middle of her actions, she asks a robot for help, saying \textit{``Could you pass that from the sofa?''} This instruction does not appear to be solvable without further information about the person's underlying intention. While such internal mental states are not directly observed, agents can infer them from human's past actions. Specifically, if the robot can observe that in previous steps, the human put several books into a box one by one, it can infer that she intends to use that box to store all the books. Based on this guess, the robot can check if there are any remaining books on the sofa and then hand them to the person.

Generally speaking, the ambiguity in the instruction mainly arises from two aspects. First, the human assumes sufficient prior knowledge about her hidden intentions~\cite{dennett1987intentional,gergely1995taking}, which is based on the common sense knowledge that people tend to group similar items together when tidying up, and the observation that the human is gathering books. Second, people make trade-offs between accuracy and efficiency of communication~\cite{grice1975logic,sperber1986relevance,clark1996using}. This leads to the challenge of building AI agents that can follow efficient, ambiguous speech that people naturally adopt when giving directions.

We consider the case where there's a human and a robot collaborating in a shared environment. The human is working on some tasks, and specifies a sub-task for the robot to help with by giving a natural language instruction. Past methods (\eg, language grounding) attempt to directly complete the specified command from the given instructions, since the human only acts as a disembodied issuer of instructions and is not another active agent in their environments. The human, as another partially observable factor in the environment, has been overlooked. In this paper, we present a new framework, \revise{Follow Instructions with Social and Embodied Reasoning (\model)}, which suggests that we should introduce the human's intention as explicit variables for the model to draw inferences about. By leveraging this structure, our framework opts to decompose the problem into \revise{two parts -- \textbf{social reasoning} and \textbf{embodied reasoning}.} 
\revise{Specifically, \textbf{social reasoning} is aimed at predicting the sub-task for which the human is asking for assistance, which can be inferred from the context of both the instruction and the observed historical actions of the person in the shared environment. After grounding the instructions into robot-understandable tasks, the robot can then do planning and interact with the environment, in a separate \textbf{embodied reasoning} phase. To further enhance the model's ability to follow ambiguous instructions, we propose to explicitly add an extra plan recognition stage, where a set of logical predicates is used to help with inferring the human's overall plan.} We implement a Transformer-based model trained in a supervised learning manner to predict \revise{specified sub-tasks (and the human's underlying plan)} at intermediate layers. This step-by-step approach distinctly differs from the more commonly employed end-to-end methods in previous works. 

Overall, the key insight of our method is that separating social and embodied reasoning by explicitly modeling the human's intentions can significantly improve performance when following ambiguous natural language instructions. To test this hypothesis, we evaluate our models on a challenging benchmark, \benchmark (HMT) \cite{wan2022handmethat}, which involves ambiguous instruction following tasks in a text-based household environment. HMT contains a large number of physical objects and valid actions in each episode, as well as an enormous human goal space. We find that these properties make HMT challenging even for the largest state-of-the-art large language models (LLMs). As a competitive baseline, we also design a Chain-of-Thought (CoT) approach to prompt GPT-4 based on our framework. 

The experimental results reveal two important findings. \revise{First,} models which separate social and embodied reasoning using the \revise{\model} framework outperform end-to-end reasoning, in both Transformer-based models and CoT-prompted LLMs, which indicates that explicitly doing intermediate reasoning about human intentions is beneficial. \revise{Second,} training \revise{small-scale} models from scratch on this task outperforms our most sophisticated CoT prompting methods for large pre-trained LLMs, indicating that pretraining and domain-specific prompts are insufficient for LLMs to perform well on the challenging \revise{social and embodied} reasoning tasks under investigation. We conduct a thorough analysis over the failure modes of GPT-4 Turbo and gain insight into why state-of-the-art LLMs are not capable of this type of social and embodied reasoning skills needed to solve the tasks. 

To summarize, the contributions of this paper are to propose the \revise{\model framework, which performs instruction following by first using social reasoning and additional context to disambiguate what the human is asking, before using embodied reasoning to decide what actions to take to complete the task. We further introduce a human plan recognition stage to enhance social reasoning abilities when tasks are particularly complex or ambiguous.} We empirically demonstrate that our \model models show \revise{64.5\%} success rate on the test set on average, achieving the state-of-the-art on \revise{HMT} benchmark. Open-source code is available at: \url{https://github.com/Simon-Wan/FISER}.

\vspace{0.5em}
\section{Related Work}
\vspace{0.5em}
\subsection{Grounded language learning}
In order for AI to be useful to people in our homes and natural environments, non-experts need to be able to communicate with AI agents using natural language.
This issue has long captured the attention of researchers~\cite{winograd1972understand, siskind1994grounding}, and the primary challenge involves mapping natural language to concrete meanings within the physical environment. 
Several studies explore language-conditioned task completion in specific environments~\cite{ALFRED20, Suglia2021embodiedbert, kojima2021continual}.
With the emergence of LLMs, many works discussed grounding language by leveraging pre-trained LLMs ~\cite{Blukis2021persistent, Nair2022R3M, Zellers2021piglet}. A prominent example is SayCan~\cite{saycan2022arxiv}, which proposed extracting the knowledge in LLMs by using them to score the likelihood that a subtask available to the robot will help complete a high-level instruction. Although the above studies may incorporate common sense reasoning about language as well as information within the physical environment, their instructions explicitly express human intentions. For example, \revise{the most ambiguous} instruction solved by SayCan is, ``\textit{I spilled my coke, can you bring me something to clean it up?}'' where the ambiguity can \revise{still} be easily resolved \revise{given that \textit{the sponge} is the only cleaning tool in the environment}. In contrast, we address the problem that realistic human instructions omit certain information for efficiency, making them much more ambiguous, and necessitating inferring human intentions to fill in the gaps. 

\subsection{Collaborative communication} 
We consider the case where the human and the robot are working in a shared environment, which is closely related to the literature on collaborative communication (\eg Two Body Problem~\cite{Jain2019TwoBP}). CerealBar \cite{suhr-etal-2019-executing}, DialFRED~\cite{Gao2022DialFREDDA} and TEACh~\cite{Padmakumar2022TEAChTE} introduce collaborative tasks where the human works as a disembodied issuer of instructions, possibly responding to robot's questions via explicit messages. 
In contrast, we consider the problem in which the AI assistant needs to consider both explicit messages in natural language and the implicit information in observed human actions. Further, we assume that instructions are not exhaustively describing the required information, but are generated based on a trade-off between informativeness and communication cost. To this end, we focus on the \benchmark~\cite{wan2022handmethat} (HMT) benchmark, that calls for the ability to consider both explicit and implicit messages when following ambiguous instructions.
The previous state-of-the-art work~\cite{cao2024enhancing} on HMT performs iterated goal inference over the goal space in symbolic representation. However, it requires hand-crafted, pre-defined structures and extensive domain knowledge, which is not applicable in real-world scenarios.

\subsection{Goal recognition}
In our method, we hope to infer the human's intentions based on the observed historical actions, which is related to goal recognition problem~\cite{10.5555/1643031.1643119,baker2007goal,Levesque2011,Meneguzzi2021ASO}. Most of the works are based on the assumption of rationale that an agent should make (approximately) optimal decisions towards the goals every step~\cite{dennett1987intentional, gergely1995taking}.
Understanding human intentions in embodied environments has also been studied in many works; for example in Watch-and-Help~\cite{puig2021watchandhelp} the AI must infer the human's goal from demonstrations, but no natural language is involved. 
Some recent works~\cite{ying2024goma, zhang2024combo} leverage LLMs to conduct goal inference based on the observed human actions or messages. 
In this work, we employ a small language model trained from scratch to undertake this part of the reasoning, since the aim is not to achieve precise goal recognition but to assist with the step-by-step social reasoning process.

\subsection{Reasoning with intermediate steps}
This work is also inspired by the research that uses intermediate steps to solve complex reasoning problems, including formal and mathematical reasoning and program synthesis~\cite{Roy2015quantities,amini2019mathqa,chiang2019semantically,Chen2020Neural,nye2022show}. Specifically, \citeauthor{nye2022show} shows that step-wise prediction method performs better than directly predicting the final outputs in program synthesis when prompting LLMs. Chain-of-Thought (CoT)~\cite{wei2022chain} thoroughly explores how generating intermediate reasoning steps improves the performances of prompting LLMs to deal with complex reasoning tasks.
In this paper, we show that social reasoning tasks benefit from the same approaches, and demonstrate that inferring human intentions is a critical component of successful human-robot collaboration.

\vspace{0.5em}
\section{\model: Follow Instructions with Social and Embodied Reasoning}
\vspace{0.5em}
\subsection{Problem Formulation}
A human-robot Markov Decision Process is described as a tuple $\langle\gS, \gA^{h,r},\gT, \gU,R^{r},\gamma,T\rangle$.
$s\in \gS$ are object-oriented states including the locations, status and type of each object and agent. $\gA^{h,r}$ is the joint action space with $\gA^h, \gA^r$ being the sets of actions available to the human and the robot, respectively. $\gT:\gS\times\gA^{h,r}\times\gS\rightarrow\{0,1\}$ is the transition function where $\gT(s,a^{h,r}, s')=1$ if and only if taking actions $a^{h,r}$ at state $s$ gives $s'$ as the next state. $\gU$ is a set of instructions that the human can give to the robot. $R^{r}:\gS\rightarrow\bR$ is a reward function for the robot, $\gamma$ is the discount factor, and $T$ is the horizon. Throughout the paper, we consider a scenario with only a single round of instruction following for the robot. In each episode, starting from an initial state $s_0$, the human begins working in the environment, and the robot is waiting. Human stops at a time step $t'\le T$, leading to a trajectory $\tau_{t'}=(s_0,a^h_0,s_1,a^h_1,\ldots,s_{t'-1},a^h_{t'-1})$ and a final state $s_{t'}$. Then the human produces a natural language instruction $u \in \gU$ that asks the robot for help. Given $\tau_{t'}, s_{t'}$ and $u$, the robot needs to interact with the environment by taking a sequence of actions $\{a_t^r\}_{t\ge t'}$ to maximize its discounted rewards $\sum_{t=t'}^{T}[\gamma^{t-t'}R^r(s_{t+1})]$. 

\begin{figure*}[t]
    \centering
    \includegraphics[width=0.95\textwidth]{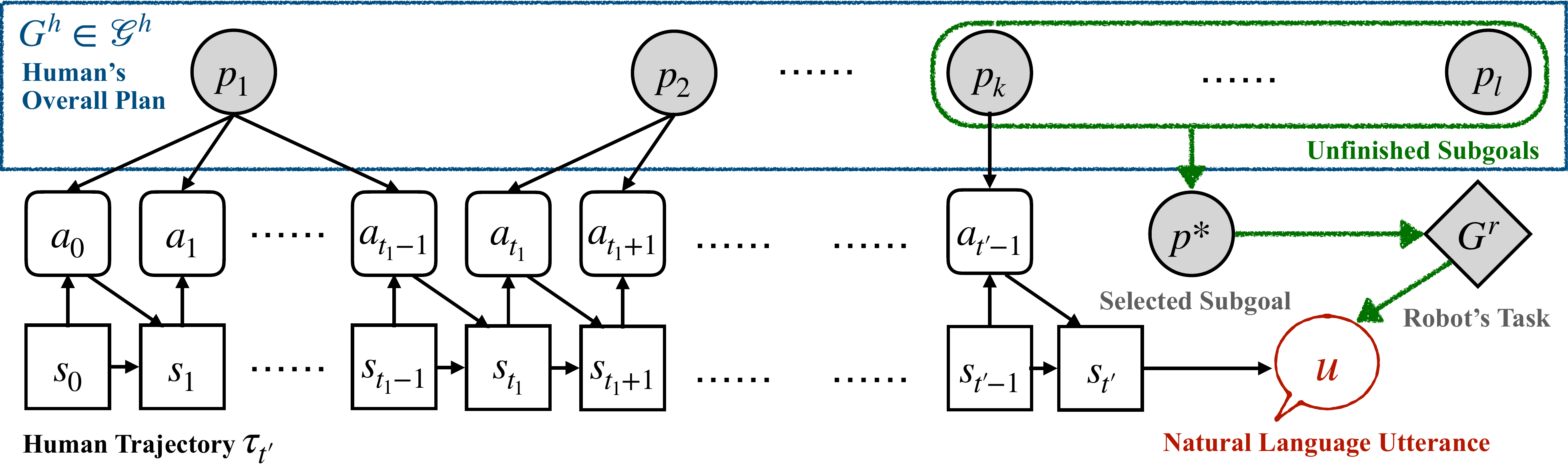}
    \caption{Graph for our problem formulation and proposed method. White nodes are observable variables, while grey nodes are unobservable. The robot is given the trajectory $\tau_{t'}$, a final state $s_{t'}$, and an utterance $u$. We propose to explicitly model human's intentions by modeling \revise{the human's overall plan} $G^h\in\mathcal{G}^h$ as a set of predicates $p_k$. We further assume that human selects a subgoal $p^*$ that needs help, and then specifies a \revise{robot's task} $G^r$, which is the underlying intention of human when saying $u$.}
    \label{fig:graph}
\end{figure*}

\subsection{Modeling the Human's Intentions}
A straightforward solution to the human-robot MDP may treat $\tau_{t'}$ and $u$ as additional state information. However, in reality, $\tau_{t'}$, $u$, and $R^r$ have important correlations: when the human is taking actions and producing instructions, their behavior can be modeled as optimizing for an internal reward function $R^h: \gS \rightarrow \R$, which is not revealed to the robot.
\revise{Our insight into this broad problem class is to leverage the causal relation between the human's behavior and their instruction by explicitly modeling hidden, unobserved variables representing the human's goals and intentions. This enables making better use of the human's trajectory to disambiguate the instruction, by recognizing the task the human is assigning to the robot.}

We start by assuming the reward functions $R^h$ can be parameterized by a set of possible goals $\gG^{h}$. The human's goal $G^h\in\gG^h$ is sampled from an underlying distribution over $\gG^h$ at the beginning of an episode, and is fixed across the horizon. However, it is not revealed to the robot directly. The goal in $\gG^h$ is usually global and complex, such as ``organize the bedroom.'' We assume the human trajectory $\tau_t$ is rational, in the sense that it was produced to maximize the reward $R^h(\cdot \mid G^h)$. Based on $G^h$ and the current progress $\tau_{t'}$, the human then 
\revise{selects a subgoal $p^*\in G^h$ (a part of the human's overall plan) for which the human would like the robot's help, such as ``having all books put in the box,'' and then specifies a task $G^r$ for the robot (\eg, asking the robot to hand over a specific book).}
The instruction $u$ is generated based on $G^r$. \revise{The relations between these variables are illustrated in \fig{fig:graph}.} 

\revise{Although we do not put specific assumptions over the structure of goals in $\gG^h$ and how $\tau_t$ is generated, we illustrate them with a simplified example in \fig{fig:graph}. We define $P$ as a set of predicates where each $p\in P$ is a classifier over states (to say whether the predicate is satisfied or not). For example, one predicate can be written as $\langle\exists y, \text{box}(y), \forall x, \text{book}(x)\Rightarrow \text{in}(x,y)\rangle$, which describes putting all books in a box.  Now we assume that the human goal $G^h = \{p_1, p_2, \cdots, p_l\}$ is a set of predicates. The human chooses to work on predicates one by one and has been working on all $p_k$'s ($1\le k\le l$) before stopping at time $t'$, and then the subgoal $p^*$ is chosen from the set of remaining predicates: $p^*\in\{p_k,\ldots,p_l\}$. Next, the human specifies a robot's task $G^r$ such that the robot actions will result in a state $s'$ where $R^h(s' \mid \{p^*\})>0$ (\ie, $G^r$ is a useful step towards $p^*$, a remaining subgoal to accomplish). Note that neither $p^*$ nor $G^r$ is accessible to the robot, since the robot can only get access to the natural language instruction $u$. For example, \textit{``could you pass that from the sofa''} could be an utterance for $G^r = \langle\text{human-holding}(\text{book\#0})\rangle$ and $p^*=\langle\exists y, \text{box}(y), \forall x, \text{book}(x)\Rightarrow \text{in}(x,y)\rangle$.}

\subsection{Step-wise Reasoning over Human Intentions}
Our model, \revise{\model,} builds on top of the factorized human-robot MDP formulation above. \revise{We formulate the problem into the social and embodied reasoning phases.}

\subsubsection{\revise{Social Reasoning: Robot's Task Recognition}}\hfill

\revise{The robot needs to disambiguate the natural language instruction $u$ into an understandable and executable task within its own goal space based on the observation of current state $s_{t'}$ and the historical trajectory $\tau_{t'}$. Therefore, we hope to estimate a function TR, such that $\text{TR}(s_{t'},\tau_{t'},u)\rightarrow G^r.$} 

\subsubsection{\revise{Social Reasoning: Human's Plan Recognition.}}\hfill

\revise{We further propose a variant that explicitly estimates the human's underlying overall plan $G^h$ based on $\tau_{t'}$ and replaces that trajectory by the predicted goal when doing instruction disambiguation. However, since recognizing the full plan is usually intractable, we opt to also take in $u$ and $s_{t'}$, and directly predict the predicate $p^*\in G^h$ (subgoal) that the human wants the robot to help with. Therefore, we learn two functions PR and TR, such that $\text{PR}(s_{t'},\tau_{t'},u)\rightarrow p^*$ and $\text{TR}(s_{t'},p^*,u)\rightarrow G^r.$}

\subsubsection{\revise{Embodied Reasoning: Grounded Planning.}}\hfill 

Once the robot goal $G^r$ is obtained, the problem is reduced to a pure grounding and planning task. We can replace the ambiguous natural language instruction $u$ by the accurately expressed robot goal $G^r$. The final grounded planning function GP should satisfy that $\text{GP}(s_{t'},\tau_{t'},G^r)\rightarrow \{a_t^r\}_{t\ge t'}$, which is basically learning a typical goal-conditioned robot policy $\pi(a|s_{y'},\tau_{t'},G^r)$\footnote{$\tau_{t'}$ should be $p^*$ instead if the PR stage is included.}.

Since the functions TR and PR involve natural language inputs, language models are required for these two modules. For GP, we can either implement planning algorithms or use neural networks, since all inputs can be symbolic.

\vspace{1em}
\section{Transformer-based Model Implementation}
\vspace{0.5em}
\label{sec:model}

\begin{figure*}[t]
    \centering\small
    \includegraphics[width=\linewidth]{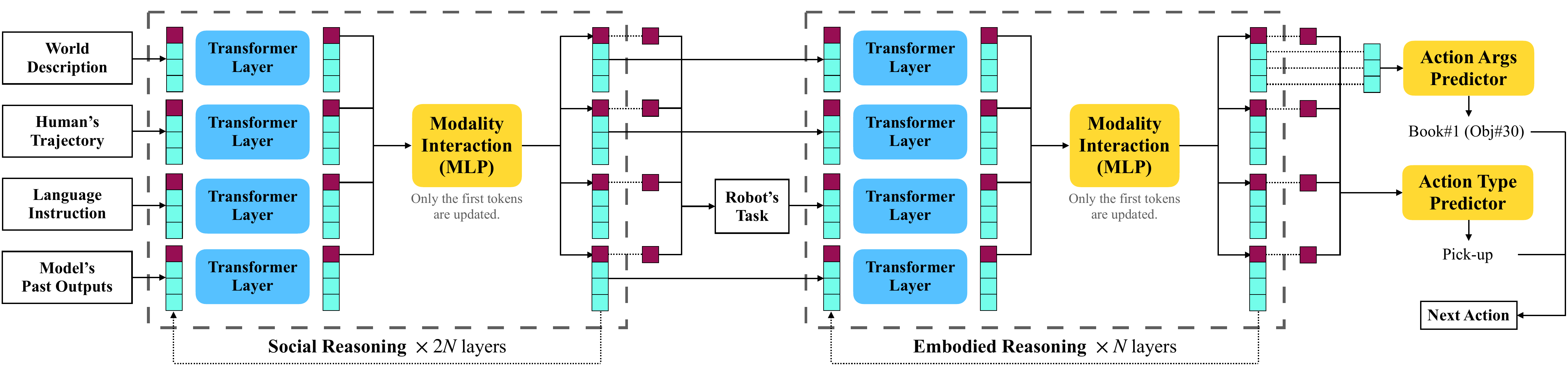}
    \caption{The Transformer-based model has four parts of inputs, which are passed separately into different Transformer Encoder Layers, and interact with each other through a Modality Interaction module after each Transformer layer. \revise{The first $2N$ layers form the social reasoning phase and the last $N$ layers form the embodied reasoning phase. The embeddings at Layer $2N$ are used for recognizing robot's task, and the last layer embeddings are used for predicting actions.}}
    \label{fig:arch}
\end{figure*}
We implement a Transformer-based model following our framework, illustrated in \fig{fig:arch}. We assume all inputs are rendered in texts, and the model needs to predict action strings.

\subsection{Inputs}
Since small-scale language models cannot process excessively long inputs, we divide the information into four parts, including world state description ($s_{t'}$), human's trajectory ($\tau_{t'}$), language instruction ($u$), and the model's past outputs. 

\subsubsection{World state description.}\hfill

We consider an object-centric representation for the world state. Specifically, the world state is described as a sequence of object tokens. For each object in the world, we fuse the information of its category (object type and genre), attributes (e.g., \textit{size, color, is-open}), and spatial relation (\textit{inside} or \textit{on top of} another object) into one single embedding, which we called as an object token. The human and the robot are also treated as two special ``objects''.

\subsubsection{Human's trajectory.}\hfill

A straightforward way to represent the human's trajectory is to directly use a paragraph of text to describe the action sequence, \eg, ``\textit{the human picks up book\#1 from the table}''. To better align it with the world state, we replace the embeddings for object names (``book\#1'') by the object tokens that we obtained in the world state description.

\subsubsection{Language instruction.} \hfill

The natural language instruction sentence is tokenized and then directly turned into embeddings.

\subsubsection{Model's past outputs.} \hfill

Every time the agent takes a step, the environment returns a sentence describing the effect of its action, \ie, the update in observations. In each episode, such sentences for past steps are concatenated and served as an extra input, \eg, ``\textit{\ldots [SEP] slice apple 0 [SEP] The apple (apple 0) is sliced into two pieces\ldots}'' The concatenated result is also tokenized and then turned into embeddings.
This information is necessary because the model needs to know what it has done and whether the world state is changed.

\subsection{Model Architecture Overview}
\label{sec:arch-overview}

The proposed model consists of $3N$ ($N=3$) encoder layers that are used to update the representation over all four parts of the inputs, in order to predict the human's intentions or the robot's actions. 
\revise{Specifically, we use the embeddings at Layer $2N$ to predict the robot's task $G^r$ (social reasoning phase). Then, we replace the instruction input for Layer $2N+1$ by the predicted $G^r$ and then use the last layer embeddings to predict robot's actions. If the Human's Plan Recognition stage is further included, we use the embeddings at Layer $N$ to predict the selected human subgoal $p^*$ and then replace the trajectory input for Layer $N+1$ by the predicted $p^*$.}

The model is trained in either a multi-staged (MS) or an end-to-end (E2E) manner. 
The E2E models are trained to directly output robot's actions, but an auxiliary loss is applied over their intermediate predictions \revise{of robot's task (and human's plan)}. The MS models, however, disentangle the social reasoning \revise{(functions TR and PR)} from embodied reasoning \revise{(function GP)}, and train them as two separate modules. The latter module is trained with the ground-truth $p^*$ or $G^r$, but is evaluated using the predictions from the former module.

\subsection{Encoder Layers}

Each encoder layer is composed of four Transformer layers and a Modality Interaction module. 

\subsubsection{Transformer Layers.}\hfill

We use four separate Transformer encoder-only layers to process the inputs. We remove the positional encoding for the world state description, because we do not expect the model to learn an ordering of objects.

\subsubsection{Modality Interaction.}\hfill

In order to fuse the information from the four parts of inputs, we design a modality interaction module (an MLP) within each layer, following the architecture proposed by GreaseLM~\cite{zhang2021greaselm}. We reserve a special ``interaction'' token at the front of each part of inputs. These tokens are expected to gather respective information in the Transformer layers and then interact with each other through this MLP. The updated special tokens will then replace the original first token in each part of inputs. 

\subsection{Prediction Layers}

Now we introduce our predictors for intermediate reasoning steps and the final actions. All the predictions are trained with cross entropy loss over corresponding supervisions.

\subsubsection{Social Reasoning: Human's Plan Recognition.} \hfill

\revise{We assume that there is a vocabulary of concepts that allow us to represent human goals as first order logic predicates (\eg, $\langle\exists y, \text{box}(y),$ $\forall x, \text{book}(x)\Rightarrow \text{in}(x,y)\rangle)$.}
While such logical predicates could be more complex, here we assume human plans follow a simpler form: a Q(uantifier), a S(ubjective), a V(erb), and an O(bjective) (\eg, $\langle \text{for-all}, \text{book}, \text{inside}, \text{box}\rangle$). Therefore, the human's plan recognition module of our model needs to predict a tuple of four tokens.
The prediction is conditioned on the embeddings of the four inputs. Specifically, we calculate the log-likelihood over all tuples as follows, where the predictions of the Subjective and Objective are conditioned on the predicted values of Quantifier and Verb:
\begin{align*}
    \log\Pr[\text{Q}, \text{S}, \text{V}, \text{O}] \approx&\log\Pr[\text{Q}]+\log\Pr[\text{V}]\\+&\log\Pr[\text{S}|\text{Q},\text{V}]+\log\Pr[\text{O}|\text{Q},\text{V}],
\end{align*}

\subsubsection{Social Reasoning: Robot's Task Recognition.} \hfill

\revise{The main target of the entire social reasoning phase is to predict the task assigned to the robot}, such as a specific object to manipulate. Note that in this step the model needs to specify a concrete object in the world, while the Subjective and Objective predictions in previous plan recognition step are object types. 

\subsubsection{Embodied Reasoning: Action Prediction.} \hfill

The model needs to output the next action in each step, which we assume to be a triple consisting of action type and one or two arguments, (\eg, move-to(sofa), put-into(book\#0, box\#1)). The action type prediction is conditioned on the embeddings of the four inputs,
while the arguments are further conditioned on the predicted action type. 
Specifically, we calculate the log-likelihood over all actions (no matter applicable or not).
\begin{align*}
    \log\Pr[\text{Action}, &\text{Arg1}, \text{Arg2}]\approx\log\Pr[\text{Action}]\\+&\log\Pr[\text{Arg1}|\text{Action}]+\log\Pr[\text{Arg2}|\text{Action}].
\end{align*}
We assume the model has access to all applicable actions at each step, so we take the maximum over all applicable triples to get the final prediction.

\vspace{0.5em}
\section{Experiments}
\vspace{0.5em}
We evaluate our framework by training a Transformer-based model from scratch for the challenging \benchmark benchmark~\cite{wan2022handmethat},
and then compare them with multiple competitive baselines, including the state-of-the-art prior work on HMT, and the CoT prompting on the largest available pre-trained language models. We will investigate the following hypotheses through empirical analysis. 

\vspace{1em}

\xhdr{\textit{H1:}} \textit{
Explicitly modeling underlying human intentions works better than directly predicting actions.
}
\begin{enumerate}
    \item[\textit{(a)}] \textit{\revise{Separating the social and embodied reasoning steps by explicitly recognizing the robot's task is beneficial.}}
    \item[\textit{(b)}] \textit{\revise{Explicitly recognizing the human's plan further helps with the social reasoning stage.}}
\end{enumerate}

\vspace{0.5em}

\xhdr{\textit{H2:}} 
\textit{Pre-trained LLMs, despite having access to common-sense knowledge, do not adequately perform the complex social and embodied reasoning in this task.  Incorporation of domain-specific knowledge through CoT can help.}

\subsection{\benchmark Environment}
We evaluate our models over the \benchmark (version 2) dataset~\citep{wan2022handmethat}. It introduces a household ambiguous instruction following task rendered in text.
\benchmark instructions are split into four difficulty levels, and the gaps between levels correspond to different challenges. The instruction in a Level 1 task has no ambiguity---it is a pure planning task. A Level 2 task requires social reasoning where a robot can successfully accomplish the task if it can also infer the goal from the human trajectory. On Level 3, the robot needs to further consider pragmatic reasoning in language use. For example, if there are multiple books everywhere in the room and only one coat, which is on the sofa, and both the book and the coat are helpful to the human's goal, the human might say ``\textit{Could you pass that from the sofa?}''. Using pragmatics, we can understand the human is asking for the book, since ``\textit{from the sofa}'' is required to disambiguate which book the human is referring to, but if the human wanted the coat they could simply ask for it.
The final Level 4 contains tasks with inherent ambiguities that cannot be resolved with the existing information, but can potentially be resolved with a strong prior over what human is likely to do. For example, taking one more fruit will complete the goal of packing picnics, but there are many kinds of fruits in the refrigerator to choose from. From the perspective of completing the goal, any fruit will do, but human preferences may make a difference--the human may want apples instead of bananas at this time. 

We evaluate all the models on their success rates in achieving the robot's goal. \revise{Note that the original evaluation metric in \benchmark additionally considers the number of robot's steps. An agent can trivially improve success rate with increased steps, by simply searching all objects in a brute force, trial-and-error fashion. We believe that enumeration over objects is not realistic in the real world, so we restrict our experiments to one trial (to be completed within 4 or 5 steps).}

\subsection{Model Details}
\label{sec:ablation}

\subsubsection{Baseline models.}\hfill

We compare our results to human performance on the task (Human), a hand-coded baseline (Heuristic) which has access to ground-truth symbolic state information, and a neural network baseline (Seq2Seq~\cite{10.5555/2969033.2969173}) introduced in the \benchmark paper. 
The existing SOTA work~\cite{cao2024enhancing} was implemented based on the original \benchmark (version 1) dataset. Therefore, to faciliate comparison with this work we additionally report the results of our \model models over version 1 data points that lie in the version 2 domain. Further details of this comparison are provided in Appendix \ref{appendix:baselines}.

\subsubsection{Transformer-based model.}\hfill

Following the proposed model architecture, we implement a set of Transformer-based models. 
We compare the implementations with no intermediate supervision (Transformer), with Robot's Task Recognition only (Transformer+FISER), and with Human's Plan Recognition in addition (Transformer+FISER+PR). Our models are trained from scratch because existing small-scale pre-trained models cannot handle the excessive token lengths of \benchmark data inputs. 
\revise{We compare two ways of training the Transformer-based model using FISER framework, end-to-end (E2E) or multi-staged (MS), as explained in Section \ref{sec:arch-overview}. This comparison aims to provide insights for whether to block the gradient flow from embodied reasoning back to the social reasoning module. 
We further report the accuracy of the intermediate prediction steps for these Transformer-based models, including QSVO (simplified human subgoal), and Obj (concrete object to be manipulated in expert demonstration, \ie, the robot's task).}
\revise{The number of parameters in an E2E model is 5.1M; the number of parameters in an MS model is 4.7M (3.0M for social reasoning and 1.7M for embodied reasoning).}

\subsubsection{Prompted GPT-4.}\hfill

We design prompting methods for GPT-4 Turbo over HMT tasks. The vanilla implementation (GPT-4) simply provides all inputs to the model and requests it to output actions. We first conduct prompt engineering (GPT-4+PE) to incorporate some domain-specific knowledge and help to parse the complex inputs. Then we implement \model framework by applying CoT prompting to explicitly predict the same intermediate data (human's plans and specified robot's tasks) that we use for Transformer-based models step-by-step, which similarly gives two models (\ie, GPT-4+FISER and GPT-4+FISER+PR). 
\revise{To be more specific, in both social reasoning steps, we prompt GPT-4 to do step-by-step reasoning except that we ask different questions. Human's Plan Recognition asks about the human's higher-level goal, while Robot's Task Recognition asks about the intended meaning of an ambiguous instruction.}
We further consider providing additional assistance by filtering out a proportion of irrelevant objects from the environment, to assess the impact of excessive item quantity on GPT-4's embodied reasoning. The prompts for GPT-4 Turbo are provided in Appendix \ref{appendix:prompts}.

\vspace{0.5em}
\section{Results}
\vspace{0.5em}
\begin{table*}[t]
\centering
\small
\addtolength{\tabcolsep}{-0.2em} 
\begin{tabular}{@{}c|ccc|cccc|ccc|cc@{}}
\toprule
\multirow{2}{*}{Model} & \multicolumn{3}{c|}{Baseline Models} & \multicolumn{4}{c|}{GPT-4 Turbo} & \multicolumn{3}{c|}{Transformer-based Models} & \multicolumn{2}{c}{On HMT Version 1} \\
                       & Human  & Heur.  & Seq2Seq & Vanilla  & +PE   & +FISER & +PR  & Vanilla       & +FISER        & +PR           & \citeauthor{cao2024enhancing}               & FISER         \\ \midrule
Level 1                & 100.0  & 100.0      & 30.4    & 72.0         & 82.0      & 80.0       & 77.0     & 77.7$\pm$1.6  & \textbf{89.0}$\pm$1.5  & 72.0$\pm$1.5  & 27.7$\pm$0.3      & 89.7$\pm$0.4     \\
Level 2                & 80.0   & 64.0       & 28.8    & 16.0     & 25.0  & 36.0   & 34.0 & 55.3$\pm$0.4  & \textbf{74.0}$\pm$0.3  & \textbf{74.0}$\pm$0.9  & 24.8$\pm$0.4      & 63.0$\pm$0.3     \\
Level 3                & 40.0   & 39.0       & 12.8    & 5.0         & 13.0      & 18.0       & 17.0     & 36.3$\pm$1.0  & \textbf{52.3}$\pm$2.3  & \textbf{52.3}$\pm$1.0  & 21.0$\pm$0.1      & 28.3$\pm$2.5     \\
Level 4                & 30.0   & 29.0       & 14.8    & 9.0         & 9.0      & 17.0       & 20.0     & 38.3$\pm$1.4  & 42.7$\pm$0.2  & \textbf{51.0}$\pm$1.2  & 21.7$\pm$0.2      & 40.7$\pm$1.9    \\
\bottomrule
\end{tabular}
\caption{Success rate (\%) of models over HMT in the fully observable setting. \revise{The results for Transformer-based models are the mean and standard error values over three runs.} *\citet{cao2024enhancing} is evaluating version 1 of \benchmark dataset, and thus we provide the results of \revise{Transformer+FISER model} over a subset of version 1 for a fair comparison. \revise{Overall, FISER improves the performance across all levels compared to the vanilla Transformer.} While applying FISER and PR to GPT-4 improves its performance, overall GPT-4 cannot perform well on these tasks even with very careful prompting, achieving less than half the success rate of our model for ambiguous instructions in levels 2-4.}
\label{tab:results}
\end{table*}
\begin{table*}[tp]
\centering\small
\begin{tabular}{l cccc}
\toprule
Model
                       & Level 1     & Level 2    & Level 3   & Level 4      \\ \midrule
End-to-End     &  77.7$\pm$1.6  (N/A, N/A)   &    55.3$\pm$0.4 (N/A, N/A)    &   36.3$\pm$1.0 (N/A, N/A)      &    38.3$\pm$1.4 (N/A, N/A) \\
End-to-End+FISER     &  74.3$\pm$0.2 (N/A, 87.8)   &    73.7$\pm$0.5 (N/A, 80.9)    &   47.3$\pm$2.1 (N/A, 73.1)      &    41.3$\pm$1.9 (N/A, 64.7) \\
End-to-End+FISER+PR &  61.0$\pm$1.5 (73.2, 81.3)   &    66.7$\pm$1.1 (64.9, 81.0)    &   47.0$\pm$1.2 (59.0, 73.4)      &    42.0$\pm$0.3 (55.0, 66.5) \\ \midrule
Multi-Staged+FISER     &  \textbf{89.0}$\pm$1.5 (N/A, 93.4)   &    \textbf{74.0}$\pm$0.3 (N/A, 82.3)    &   \textbf{52.3}$\pm$2.3 (N/A, 76.5)      &    42.7$\pm$0.2 (N/A, 68.1) \\
Multi-Staged+FISER+PR  &  72.0$\pm$1.5 (74.4, 82.4)   &    \textbf{74.0}$\pm$0.9 (71.9, 82.3)    &   \textbf{52.3}$\pm$1.0 (67.1, 75.6)      &    \textbf{51.0}$\pm$1.2 (65.3, 71.2) \\
\bottomrule
\end{tabular}
\caption{Comparison between End-to-End and Multi-staged training of Transformer-based models over HMT in the fully observable setting. All models are evaluated by the success rate (\%). The prediction accuracy (\%) of QSVO and Obj (intermediate outputs) are presented in parentheses. \revise{The results are the mean values over three runs, and the standard error values for success rates are provided. Overall, training in a multi-staged manner works better than end-to-end in our tasks, implying that fully separating social from embodied reasoning provides the best performance.}}
\label{tab:ablation}
\end{table*}

We evaluate all the models over the \benchmark (version 2) dataset in the fully observable setting. Overall, our best-performing \revise{Transformer+FISER} model achieves a \revise{64.5\%} success rate on average, achieving the state-of-the-art on the \benchmark benchmark.
The main results are presented in Table \ref{tab:results}.
Now we discuss the previously stated hypotheses.

\paragraph{H1: Explicitly modeling human intentions works better than directly predicting actions. } \hfill

\subsubsection{\revise{(a) Separating the social and embodied reasoning steps by explicitly recognizing the robot's task is beneficial.}}\hfill

For both prompted GPT-4 Turbo and Transformer-based models, explicitly predicting the robot's task significantly improves the success rates across all difficulty levels, which supports our hypothesis that separating the social and embodied reasoning steps is beneficial in these complex reasoning tasks. The comparison between two different training schemes of our Transformer-based models are presented in Table \ref{tab:ablation}. Results show that training in a multi-staged manner works better than end-to-end in our tasks. It may imply that the low-level grounded planning (embodied reasoning) is requiring a sufficiently different task representation from inferring human's internal goals (social reasoning), that allowing gradients from the embodied reasoning module to flow into the social reasoning module actually hurts performance. It is a further support on empirical side that we should make explicit inferences about human intentions as intermediate reasoning steps.

\subsubsection{\revise{(b) Explicitly recognizing the human's plan further helps with the social reasoning stage.}}\hfill

When we further include the Human's Plan Recognition (PR) stage, we find that it only helps for the most ambiguous cases (like in Level 4). For GPT-4 Turbo, adding PR is showing approximately the same performances as normal FISER method.  
We attribute this to the fact that pre-trained LLMs are not good at leveraging hierarchical predictions to improve on this task. 
\revise{For Transformer-based models, introducing PR gives better performance on Level 4, but is harmful to the simplest Level 1.}  We attribute the poor performance in Level 1 to the fact that such simple tasks do not require knowing the humans' high-level goal. Therefore, forcing the model to predict this information reduces the model's capacity to focus on planning for low-level actions. 
On the other hand, the improved performance on Level 4 shows that explicit human's plan recognition helps to better learn priors over human intentions. Even on these intrinsically ambiguous tasks, the model can leverage the strong prior to take helpful actions.

\paragraph{H2: Pre-trained LLMs, despite having access to common-sense knowledge, do not adequately perform the complex social and embodied reasoning in this task.  Incorporation of domain-specific knowledge through CoT can help.}\hfill

Results show that training \revise{much smaller, more efficient} Transformer-based models from scratch is exhibiting \revise{about 70\%} increased performance than prompting state-of-the-art pre-trained LLMs. GPT-4 Turbo's results on Level 1 show it has the capability to do some level of embodied reasoning when given explicit tasks. However, the performance drop on subsequent levels indicates that the required knowledge to solve \benchmark tasks is not fully covered by common-sense knowledge in pre-trained LLMs. With well-designed prompt engineering (PE) that contains some domain knowledge (\eg, goal space and \revise{few-shot examples}), GPT-4 Turbo improves significantly across all difficulty levels. \revise{However, even with careful CoT prompts and few-shot examples, it is far from small-scale Transformer models across all levels.} 

\begin{figure}[ht]
    \centering\includegraphics[width=0.9\linewidth]{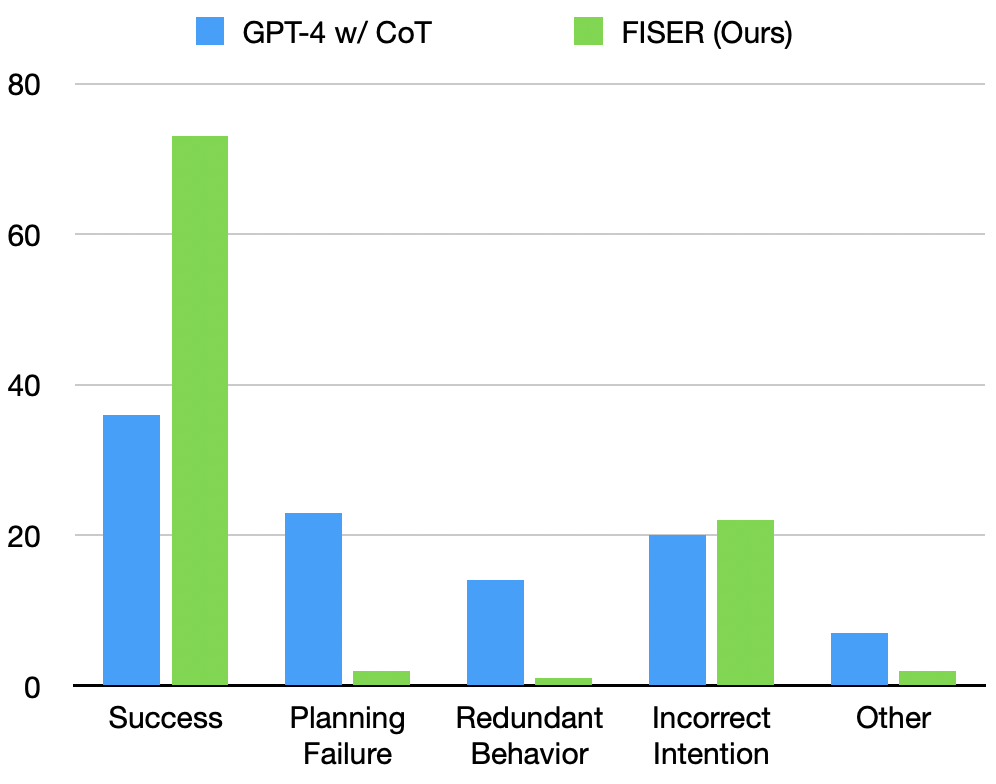}
    \caption{Failure case analysis for state-of-the-art LLMs with CoT prompts following FISER framework versus Transformer-based models trained from scratch with the FISER framework over 100 data points on Level 2.}
    \label{fig:failure}
    \vspace{-1em}
\end{figure}

As a qualitative analysis, we evaluate the failure cases for FISER framework on prompting GPT-4 Turbo and training Transformer-based models. We provide the results on Level 2 over 100 data points in \fig{fig:failure}. We define the failure modes as follows:
\begin{itemize}
    \item \textbf{Planning Failure:} Hallucination (go to a place where the target object is not located at, i.e., fail to locate the object), missing steps, or invalid actions.
    \item \textbf{Redundant Behavior:} The model gives the human an object that is already at its target location or even one which was just manipulated by the human.
    \item \textbf{Incorrect Intention:} For GPT-4, common-sense reasoning is performed but not aligned with the ground-truth human intention. (As an example, it will pick up radish given that the person was preparing soup, because "radish is a component of some soups." But this relationship does not exist in the human's goal space.) For Transformer, the model can reach the object it wants, but that object is not what the human wants.
\end{itemize}
Here, Planning Failure corresponds to the difficulty in embodied reasoning, while Incorrect Intention and Redundant Behavior corresponds to the challenges in social reasoning.\footnote{A single failure case could result from multiple errors, which means it is possible that a failure case that counts as Planning Failure could also apply to Incorrect Intention.} Results show that a Transformer-based model can effectively avoid both Planning Failure and Redundant Behavior, while recognizing the target object correctly is still a challenging issue. Note that the Planning Failure rate of GPT-4 Turbo is around 20\%, which aligns with its performance on Level 1 (80\% success rate) in Table \ref{tab:results}, since Level 1 only incorporates challenges in embodied reasoning. 
\revise{We believe the unsatisfactory performance of GPT-4 Turbo is because the prompting methods alone cannot provide the model with the type of social and embodied reasoning needed to solve this task. The information learned from language datasets collected online may also be significantly different from that required for this household assistance task. 
Training a small-scale model, however, can solve the problem more efficiently and reliably.}

\revise{We conducted an additional experiment to see if the performance of the pre-trained LLMs could be improved. 
Here we provide additional assistance by filtering out a proportion of irrelevant objects from the environment (which assumes access to the ground-truth human goals), to assess the impact of excessive item quantity on GPT-4’s embodied reasoning.
\begin{figure}[ht]
    \centering
    \includegraphics[width=\linewidth]{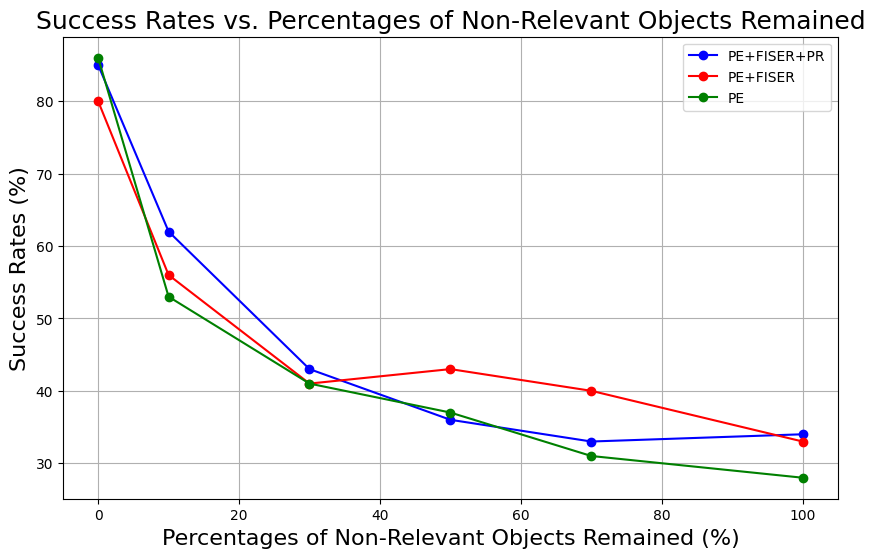}
    \caption{Success rates of GPT-4 Turbo under different settings given that different percentages of irrelevant objects are filtered out from the environment.}
    \label{fig:filter}
\end{figure}
The results are presented in \fig{fig:filter}. Even in the case that only relevant objects are remained, GPT-4 Turbo can only achieve less than 85\% success rate, which aligns with the planning failure rates in the failure case analysis above, indicating that the LLM cannot easily resolve the difficulty in embodied reasoning. As the irrelevant objects increase, the success rate of GPT-4 Turbo drops dramatically, showing the challenges of social reasoning in HandMeThat tasks.
To summarize, we observe that LLM's performance relies on a very large proportion of objects being filtered out, which provides further insight that LLMs cannot effectively select relevant environment information and focus on relevant objects, which is required in embodied reasoning.}

\vspace{1em}
\section{Conclusion}
\vspace{0.5em}
We study the challenging HandMeThat benchmark, comprising ambiguous instruction following tasks requiring sophisticated embodied and social reasoning. We find that existing approaches for training models end-to-end, or for prompting powerful pre-trained LLMs, are both insufficient to solve these tasks. We hypothesized that performance could be improved by building a model that explicitly performs social reasoning to infer the human's intentions from their prior actions in the environment. Our results provide evidence for this hypothesis, and show that our approach, \revise{Follow Instructions with Social and Embodied Reasoning (\model)},  enhances performance over the most competitive prompting baselines by \revise{70\%}, setting the new state-of-the-art for HandMeThat.

\appendix
\vspace{1em}
\bibliography{reference_aaai}
\newpage
\section{Model Implementation Details}
\label{appendix:models}
\subsection{Baseline Models}
\label{appendix:baselines}

\subsubsection{Baselines in \benchmark}\hfill

\benchmark paper provides a series of simple baselines to compare with, including Human, Random, Heuristic, Seq2Seq, and DRRN. We use their reported results on \benchmark Version 2 dataset. 

\revise{As introduced in the benchmark paper, the Heuristic model is an agent that heuristically repeats the previous human actions. To be more specific, this model has access to all object states and the underlying logic formula of the utterance. Therefore, it is only applicable in the fully-observable setting. Upon receiving the instruction, the agent generates all possible groundings of instruction and then compares them to the observed human trajectory. The key heuristic of this model is that: human tends to quest for objects that are in the same categories as the previously manipulated ones. The Heuristic model guesses the grounding of the objects in the utterance based on this heuristic. }

\subsubsection{\citet{cao2024enhancing}}\hfill

The previous SOTA work is trained and evaluated on Version 1 dataset. Their code for implementing the method on \benchmark was not released until this paper is submitted. We provide the results of our FISER model over a subset of version 1. Here the subset is selected in a way that the instruction type should be ``bring me'', and the goal should be included in Version 2 dataset. We argue it is a fair comparison for the following reasons:
\begin{enumerate}
    \item They build in the goal space to their model with hand-crafted ways, while we are using a simplified version of the goal as supervision to train the models.
    \item They use rule-based information extractor to deal with long and redundant text inputs, while we are using symbolic representation for the world state description.
    \item They claim that they are doing partially observable settings, but they are actually having full access to all the objects in the environment at least during the training phase, and didn't learn to navigate in the environment. 
    \item They only filtered a small set of goals among all 69 goal templates of Version 1 dataset. We, however, are training on 25 different possible goal templates, and evaluate our models on Version 1 data where the goal is seen but not the data.
\end{enumerate}

\subsection{Transformer-based Models}
The model architecture and implementation details are discussed in the main text. We will release the codebase for our models in the future.

\subsubsection{Hyperparameters}\hfill

\revise{The number of trainable parameters for an end-to-end model is 5,057,945; the numbers of trainable parameters for social reasoning and embodied reasoning modules within a multi-staged model are 3,020,483 and 1,701,750 respectively. We train the end-to-end model and the embodied reasoning module for 20 epochs, and train the social reasoning module for 40 epochs. The number of training epochs are chosen based on the average performance on validate sets. We choose $32$ as batch size by sweeping over $32, 64, 128$. We choose Adam with learning rate $1e-4$, betas $(0.9, 0.98)$, and epsilon $1e-9$. We choose $0, 1, 2$ as the random seeds and train all the models for three runs.}

\subsubsection{Compute Resources}\hfill

\revise{We use NVIDIA RTX A6000 to train our models. Each model is trained on a single GPU. The end-to-end model and embodied reasoning module take around 8 hours for each training epoch, while the social reasoning module takes less than our hour each epoch. Given that the embodied reasoning module is trained for 20 epochs, it takes approximately a week to train.}

\subsection{Prompted GPT-4}
We used GPT-4 Turbo from OpenAI API. GPT-4 is a large multimodal model that can solve difficult problems with greater accuracy than OpenAI's previous models. It has broader general knowledge and advanced reasoning capabilities, which makes GPT-4 Turbo suitable for our complex task that involves social reasoning and world assumptions. We paid to get text generation from the API.

\newpage
\onecolumn
\section{GPT-4 Prompts}
\label{appendix:prompts}
There are four rungs of system prompts for GPT-4, each of them building on the previous rung: GPT-4, GPT-4+PE, GPT-4+PE+FISER, GPT-4+PE+FISER+PR. We list the prompts as follows. Each prompt is split into multiple parts, and each part is put in one box. We omit the boxes that have shown up in previous prompts. The data point is placed at the word ``\textcolor{red}{Data}''.
\subsection{GPT-4}

\xhdr{Part 1: Task information.}

\xhdr{Part 2: Current data point.}

\noindent\fbox{%
    \parbox{\textwidth}{%
You are a robot. Every time you will receive a description of the world and an instruction given by the human in the following templates:

[World Description]: 

[Human Trajectory]: 

[Human Instruction]:

You are expected to generate a sequence of actions that are consistent with the human's instruction and the world description.

Please remember that each object has a number ID.

!!!!!THE LAST PART OF THE RESPONSE should be a sequence of actions starting with the string "Actions:" (!!Only actions! No numbering for actions, no description, no extra words)!!!!!

The available action templates contain:

move to XXX

pick up XXX

pick up XXX from XXX

put XXX into XXX

put XXX onto XXX

take XXX from XXX

give XXX to human

open XXX

A typical answer format is as follows, where the "open" step is optional.

Actions:

move to XXX

open XXX

pick up XXX

move to human

give XXX to human
    }%
}

\noindent\fbox{%
    \parbox{\textwidth}{%
======================== Now please begin to generate the output for given scenarios ========================

[World Description]: \textcolor{red}{Data}

[Human Trajectory]: \textcolor{red}{Data}

[Human Instruction]: \textcolor{red}{Data}

You are expected to generate a sequence of actions that are consistent with the human's instruction and the world description.

Please remember that each object has a number ID.

!!!!!THE LAST PART OF THE RESPONSE should be a sequence of actions starting with the string "Actions:" (!!Only actions! No numbering for actions, no description, no extra words)!!!!!

======================== Let's start! ========================
    }%
}

\subsection{GPT-4+PE}

\xhdr{Part 1: Task information.} Same as GPT-4, omitted.

\xhdr{Part 2: Domain specific knowledge.} 

\xhdr{Part 3: Data point example.} 

\xhdr{Part 4: Answer example.} 

\xhdr{Part 5: Current data point.} Same as GPT-4, omitted.

\noindent\fbox{%
    \parbox{\textwidth}{%

NOTE!!: Remember to open something like refrigerator if it is closed before pick up something. 
There are some openable objects: vessel, bag, box, package, cabinet, microwave, oven, dishwasher, refrigerator

Take an action like open refrigerator, open cabinet, open box, open bag, open vessel, open package, open microwave, open oven, open dishwasher

========================================================================================

To let you answer better, we give you the description of the goal space for the world as follows. The human has one of the following goals:

Goal 0: boxing books up for storage

Goal description: Put all the paper products of a certain kind into a box.

Goal 1: bring in wood

Goal description: Put all the building materials of a certain kind on top of the floor.

Goal 2: clearing the table after dinner

Goal description: Put all of one certain kind of cutlery inside some bucket, put all of another kind of cutlery into another bucket, and put all the flavorer of a certain kind into a bucket.

Goal 3: collect misplaced items

Goal description: Put all of a certain footwear, a certain decoration, and a certain paper product on top of the table.

Goal 4: collect aluminum cans

Goal description: Put all drinks into the ashcan.

Goal 5: installing alarms

Goal description: Put an electrical device on the table, on the countertop, and on the sofa.

Goal 6: laying tile floors

Goal description: Put all of a certain building material on top of the floor.

Goal 7: loading the dishwasher

Goal description: Put all the tablewares of two certain kinds and all of a certain vessel into the sink.

Goal 8: moving boxes to storage

Goal description: Put all the boxes on top of the floor.

Goal 9: oraganizing boxes in garage

Goal description: Put all of a certain plaything into a box, put all of a certain cutlery into another box, put all of a certain cleasing thing into a third box, and put all the boxes on top of the floor.

Goal 10: organize file cabinet

Goal description: Put all the writing tools of a certain kind on top of the table, and put all the paper products of a certain kind into the cabinet.

Goal 11: pick up trash

Goal description: Put all of a certain paper product and all of a certain drink into the ashcan.

Goal 12: put away Christmas decorations

Goal description: Put all the decorations of three certain kinds into the cabinet.

Goal 13: put away Halloween decorations

Goal description: Put all the vegetable of a certain kind and all the illumination tools of a certain kind into the cabinet, and put all of a certain vessels on top of the table.

Goal 14: put away toys

Goal description: Put all the playthings of a certain kind into a closed box.

Goal 15: put dishes away after cleaning

Goal description: Put all the tablewares of a certain kind into the cabinet.

Goal 16: put leftovers away

Goal description: Put all of a certain prepared food and all of a certain flavorer into the refrigerator.

Goal 17: put up Christmas decorations inside

Goal description: Put all of a certain illumination tool and all the decorations of two certain kinds on top of the table, and put all the decorations of another certain kind on top of the sofa.

Goal 18: re-shelve library books

Goal description: Put all of a certain paper product on top of the shelf.

Goal 19: serve hors d’oeuvres

Goal description: Put all of a certain baked food, a certain vegetable, a certain prepared food, and the trays on top of the table.

Goal 20: sort books

Goal description: Put all the paper products of two certain kinds on top of the shelf.

Goal 21: store food

Goal description: Put all of a certain prepared food, a certain snacks, and two certain kinds of flavorers into the cabinet.

Goal 22: store the groceries

Goal description: Put all of a certain fruit, a certain protein, and two certain kinds of vegetables into the refrigerator.

Goal 23: thaw frozen food

Goal description: Put all of a certain vegetable, a certain fruit, and a certain protein into the sink.

Goal 24: throw away leftovers

Goal description: Put all the snacks of a certain kind into the ashcan.

=========================================================================

    }%
}

\noindent\fbox{%
    \parbox{\textwidth}{%
The following is an example of prompt inputs and ground truth outputs. You should generate output in a similar way:

[World Description]: 

Welcome to the world!

In the room there is the human a countertop a sofa a bed a stove a table a shelf a toilet a cabinet a bathtub a microwave an oven a dishwasher a refrigerator a sink a pool.

Now you are standing on the floor. You are at the floor.

You see floor. On floor you can see  green highchair 0  red dusty seat 0  blue dusty seat 1  blue chair 0  red dusty chair 1  green dusty chair 2  dusty bucket 0  bucket 1  dusty bucket 2  green large closed package 2  small dusty ashcan 0  small dusty ashcan 1  large ashcan 2  xmas\_stocking 0  dusty xmas\_tree 0  xmas\_tree 1  xmas\_tree 2  shoe 0  dusty sock 0  sandal 2. In bucket 1 you can see  dishtowel 1  detergent 0. 

You see countertop. On countertop you can see  green closed bottle 0  red closed kettle 1  red closed dusty kettle 2  small dusty plate 0  red closed dusty briefcase 1  large dusty tray 2  xmas\_stocking 2  strawberry 1  strawberry 2  carrot 1  beer 1  fish 0  fish 1  parsley 1  bread 2  cookie 0  saw 0  saw 1  hammer 0  carving\_knife 0  toggled-off facsimile 0  calculator 1  mouse 0  earphone 0  pencil 1  highlighter 1  fork 1  spoon 0. 

You see dusty stained sofa. On sofa you can see  green closed duffel\_bag 0  green basket 0  green basket 1  red large open dusty box 0  red large closed dusty box 1  blue small closed dusty package 0  ribbon 1  dusty bow 1  document 1  dusty document 2  hardback 1  hat 0  cube 0. In box 0 you can see  document 0. In package 0 you can see  dusty ribbon 0  bow 0  bracelet 2  hat 2. 
You see dusty bed. On bed you can see  blue basket 2  bracelet 0  dusty jewelery 0  hardback 0  dusty apparel 0  apparel 1  ball 0. 

You see toggled-off stove. On stove you can see nothing. 

You see dusty table. On table you can see  green closed bottle 1  green closed sack 0  blue closed briefcase 0  blue closed dusty briefcase 2  small tray 1  blue large closed dusty box 2  xmas\_stocking 1  strawberry 0  banana 2  pop 0  vegetable\_oil 0  vegetable\_oil 1  sugar 1  cracker 0  cracker 1  bread 0  bread 1  cookie 2  salad 0  soup 0  pasta 2  carving\_knife 1  scraper 0  scraper 1  screwdriver 1  toggled-off printer 0  toggled-off printer 1  calculator 0  mouse 2  pencil 2  highlighter 0  lamp 0  lamp 1. On tray 1 you can see  sandwich 1  soup 2. 

You see stained shelf. On shelf you can see  blue small closed dusty package 1  saw 2  hammer 1  screwdriver 0  toggled-off scanner 0  mouse 1  earphone 1  lamp 2. 

You see dusty stained toilet. On toilet you can see  toothpaste 1  toothpaste 2. 

You see open dusty cabinet. In cabinet you can see  blue closed dusty kettle 0  small dusty bowl 0  large bowl 1  pan 2  pop 1  beer 2  juice 1  parsley 0  vegetable\_oil 2  hammer 2  scraper 2  toggled-off modem 0  toggled-off modem 1  pencil 0  dishtowel 0  rag 1  soap 1  toothpaste 0  detergent 2  bracelet 1  painting 0  dusty painting 1  painting 2  book 0  book 1  notebook 0  dusty sandal 0  sandal 1  dusty hat 1  cube 1. 

You see stained bathtub. In bathtub you can see  makeup 1  rag 0  vacuum 0  broom 0  broom 1  shampoo 2. 

You see closed toggled-off stained microwave. In microwave you can see  cooked salad 1  cooked soup 1. 

You see closed toggled-off dusty oven. In oven you can see  cooked cracker 2  cooked cake 0. 

You see closed toggled-off dishwasher. In dishwasher you can see  dusty pan 0  knife 0  spoon 1. 

You see closed toggled-on dusty refrigerator. In refrigerator you can see  large dusty mug 0  small bowl 2  large tray 0  frozen banana 0  frozen banana 1  frozen carrot 0  frozen carrot 2  frozen beer 0  frozen juice 0  frozen fish 2  frozen sugar 0  frozen tea\_bag 0  frozen tea\_bag 1  frozen cookie 1  frozen cake 2  frozen sandwich 0  frozen sandwich 2  frozen salad 2  frozen pasta 0  fork 0  knife 1. On tray 0 you can see  cake 1  pasta 1. 

You see toggled-off stained sink. In sink you can see  dusty pan 1  makeup 0  makeup 2  rag 2  hand\_towel 0  vacuum 1  scrubbrush 0. 

You see stained pool. In pool you can see  soap 0  detergent 1  shampoo 0  shampoo 1  dusty tile 0  tile 1. 

Human is currently holding nothing. Now it is your turn to help human to achieve the goal!

[Human Trajectory]: 

The human agent has taken a list of actions towards a goal, which includes:

Human moves to the cabinet.

Human opens the cabinet.

Human picks up the document 0 at the cabinet.

Human moves to the sofa.

Human opens the box 0 at the sofa.

Human puts the document 0 into the box 0.

Human stops and says, 

[Human Instruction]: 

'Please give me the dusty one.'
    }%
}

\noindent\fbox{%
    \parbox{\textwidth}{%

--------------------------------

Actions:

move to sofa

pick up document 2

move to human

give dusty document 2 to human

    }%
}

\subsection{GPT-4+PE+FISER}

\xhdr{Part 1: Task information.} Slightly different from GPT-4+PE, marked in blue.

\xhdr{Part 2: Domain specific knowledge.} Same as GPT-4+PE, omitted.

\xhdr{Part 3: Data point example.} Same as GPT-4+PE, omitted.

\xhdr{Part 4: Answer example.} 

\xhdr{Part 5: Current data point.} Slightly different from GPT-4+PE, marked in blue.

\noindent\fbox{%
    \parbox{\textwidth}{%
You are a robot. Every time you will receive a description of the world and an instruction given by the human in the following templates:

[World Description]: 

[Human Trajectory]: 

[Human Instruction]:

You are expected to generate a sequence of actions that are consistent with the human's instruction and the world description.

Please remember that each object has a number ID.

\textcolor{blue}{Remember in the final output, the type of object should be specific!! you CAN NOT say something like e.g., note !!!!!}

\textcolor{blue}{Please first think step-by-step!}

\textcolor{blue}{1. Which object do you think the human is asking for?}

\textcolor{blue}{2. What are your actions?}

\textcolor{blue}{Answer the above questions one by one NO MATTER YOU KNOW THE GOAL SPACE OR NOT!}

!!!!!THE LAST PART OF THE RESPONSE should be a sequence of actions starting with the string "Actions:" (!!Only actions! No numbering for actions, no description, no extra words)!!!!!

The available action templates contain:

move to XXX

pick up XXX

pick up XXX from XXX

put XXX into XXX

put XXX onto XXX

take XXX from XXX

give XXX to human

open XXX

A typical answer format is as follows, where the "open" step is optional.

Actions:

move to XXX

open XXX

pick up XXX

move to human

give XXX to human
    }%
}

\noindent\fbox{%
    \parbox{\textwidth}{%

[Questions]:

1. Which object do you think the human is asking for?

The human has taken some actions related to the document, and is probably asking for another document to put into the box. The human also requires that the object should be dusty. Note that there is a dusty document 2 on the sofa, so the human is probably asking for document 2.

2. What are your actions?

I should navigate to the sofa and pick up the dusty document 2. Then I should give it to the human.

Actions:

move to sofa

pick up document 2

move to human

give document 2 to human
    }%
}

\noindent\fbox{%
    \parbox{\textwidth}{%
======================== Now please begin to generate the output for given scenarios ========================

[World Description]: \textcolor{red}{Data}

[Human Trajectory]: \textcolor{red}{Data}

[Human Instruction]: \textcolor{red}{Data}

You are expected to generate a sequence of actions that are consistent with the human's instruction and the world description.

Please remember that each object has a number ID.

\textcolor{blue}{Remember in the final output, the type of object should be specific!! you CAN NOT say something like e.g., note !!!!!}

\textcolor{blue}{Please first think step-by-step!}

\textcolor{blue}{1. Which object do you think the human is asking for?}

\textcolor{blue}{2. What are your actions?}

\textcolor{blue}{Answer the above questions one by one NO MATTER YOU KNOW THE GOAL SPACE OR NOT!}

!!!!!THE LAST PART OF THE RESPONSE should be a sequence of actions starting with the string "Actions:" (!!Only actions! No numbering for actions, no description, no extra words)!!!!!

======================== Let's start! ========================
    }%
}

\subsection{GPT-4+PE+FISER+PR}

\xhdr{Part 1: Task information.} Slightly different from GPT-4+PE+FISER, marked in blue.

\xhdr{Part 2: Domain specific knowledge.} Same as GPT-4+PE+FISER, omitted.

\xhdr{Part 3: Data point example.} Same as GPT-4+PE+FISER, omitted.

\xhdr{Part 4: Answer example.} 

\xhdr{Part 5: Current data point.} Slightly different from GPT-4+PE+FISER, marked in blue.

\noindent\fbox{%
    \parbox{\textwidth}{%
You are a robot. Every time you will receive a description of the world and an instruction given by the human in the following templates:

[World Description]: 

[Human Trajectory]: 

[Human Instruction]:

You are expected to generate a sequence of actions that are consistent with the human's instruction and the world description.

Please remember that each object has a number ID.

Remember in the final output, the type of object should be specific!! you CAN NOT say something like e.g., note !!!!!

Please first think step-by-step!

\textcolor{blue}{1. What is human doing?  (Please find the most possible goal name and goal id if you are given the goal space, otherwise you should infer by yourself, but you STILL need output!)}

\textcolor{blue}{2.} Which object do you think the human is asking for?

\textcolor{blue}{3.} What are your actions?

Answer the above questions one by one NO MATTER YOU KNOW THE GOAL SPACE OR NOT!

!!!!!THE LAST PART OF THE RESPONSE should be a sequence of actions starting with the string "Actions:" (!!Only actions! No numbering for actions, no description, no extra words)!!!!!

The available action templates contain:

move to XXX

pick up XXX

pick up XXX from XXX

put XXX into XXX

put XXX onto XXX

take XXX from XXX

give XXX to human

open XXX

A typical answer format is as follows, where the "open" step is optional.

Actions:

move to XXX

open XXX

pick up XXX

move to human

give XXX to human
    }%
}

\noindent\fbox{%
    \parbox{\textwidth}{%

[Questions]:

1. What is human doing?

The human has taken some actions related to the document. By putting document 0 into box 0 on the sofa, the human is probably organizing the documents. It aligns with Goal 0: boxing books up for storage.

2. Which object do you think the human is asking for?

Given that the overall goal can be organizing the documents, the human is probably asking for another document to put into the box. The human also requires that the object should be dusty. Note that there is a dusty document 2 on the sofa, so the human is probably asking for document 2.

3. What are your actions?

I should navigate to the sofa and pick up the dusty document 2. Then I should give it to the human.

Actions:

move to sofa

pick up document 2

move to human

give document 2 to human
    }%
}

\noindent\fbox{%
    \parbox{\textwidth}{%
======================== Now please begin to generate the output for given scenarios ========================

[World Description]: \textcolor{red}{Data}

[Human Trajectory]: \textcolor{red}{Data}

[Human Instruction]: \textcolor{red}{Data}

You are expected to generate a sequence of actions that are consistent with the human's instruction and the world description.

Please remember that each object has a number ID.

Remember in the final output, the type of object should be specific!! you CAN NOT say something like e.g., note !!!!!

Please first think step-by-step!

\textcolor{blue}{1. What is human doing?  (Please find the most possible goal name and goal id if you are given the goal space, otherwise you should infer by yourself, but you STILL need output!)}

\textcolor{blue}{2.} Which object do you think the human is asking for?

\textcolor{blue}{3.} What are your actions?

Answer the above questions one by one NO MATTER YOU KNOW THE GOAL SPACE OR NOT!

!!!!!THE LAST PART OF THE RESPONSE should be a sequence of actions starting with the string "Actions:" (!!Only actions! No numbering for actions, no description, no extra words)!!!!!

======================== Let's start! ========================
    }%
}

\end{document}